\documentclass[10pt,twocolumn,letterpaper]{article}

\usepackage{cvpr}
\usepackage{times}
\usepackage{epsfig}
\usepackage{graphicx}
\usepackage{amsmath}
\usepackage{amssymb}
\usepackage{isomath,commath,stmaryrd,bm,booktabs,subfig,multirow}

\usepackage[pagebackref=true,breaklinks=true,letterpaper=true,colorlinks,bookmarks=false]{hyperref}

 \cvprfinalcopy 

\def\cvprPaperID{5245} 

\ifcvprfinal\fi
\begin{document}

\title{Exploiting Edge Features in Graph Neural Networks}

\author{
  Liyu Gong\textsuperscript{1}, \quad Qiang Cheng\textsuperscript{1,2}\\
  \textsuperscript{1} Institute for Biomedical Informatics, University of Kentucky, Lexington, USA\\
  \textsuperscript{2} Department of Computer Science, University of Kentucky, Lexington, USA\\
  {\tt\small \{liyu.gong, Qiang.Cheng\}@uky.edu}
}

\maketitle

\begin{abstract}
  Edge features contain important information about graphs. However,
  current state-of-the-art neural network models designed for graph
  learning, e.g. graph convolutional networks (GCN) and graph
  attention networks (GAT), adequately utilize edge features,
  especially multi-dimensional edge features. In this paper, we build
  a new framework for a family of new graph neural network models that
  can more sufficiently exploit edge features, including those of
  undirected or multi-dimensional edges. The proposed framework can
  consolidate current graph neural network models; e.g. graph
  convolutional networks (GCN) and graph attention networks (GAT). The
  proposed framework and new models have the following novelties:
  First, we propose to use doubly stochastic normalization of graph
  edge features instead of the commonly used row or symmetric
  normalization approches used in current graph neural
  networks. Second, we construct new formulas for the operations in
  each individual layer so that they can handle multi-dimensional edge
  features. Third, for the proposed new framework, edge features are
  adaptive across network layers. As a result, our proposed new
  framework and new models can exploit a rich source of graph
  information. We apply our new models to graph node classification on
  several citation networks, whole graph classification, and
  regression on several molecular datasets. Compared with the current
  state-of-the-art methods, i.e. GCNs and GAT, our models obtain
  better performance, which testify to the importance of exploiting
  edge features in graph neural networks.
\end{abstract}

\section{Introduction}
\label{sec:introduction}

\begin{figure}[t]
  \centering
  \includegraphics[scale=0.5]{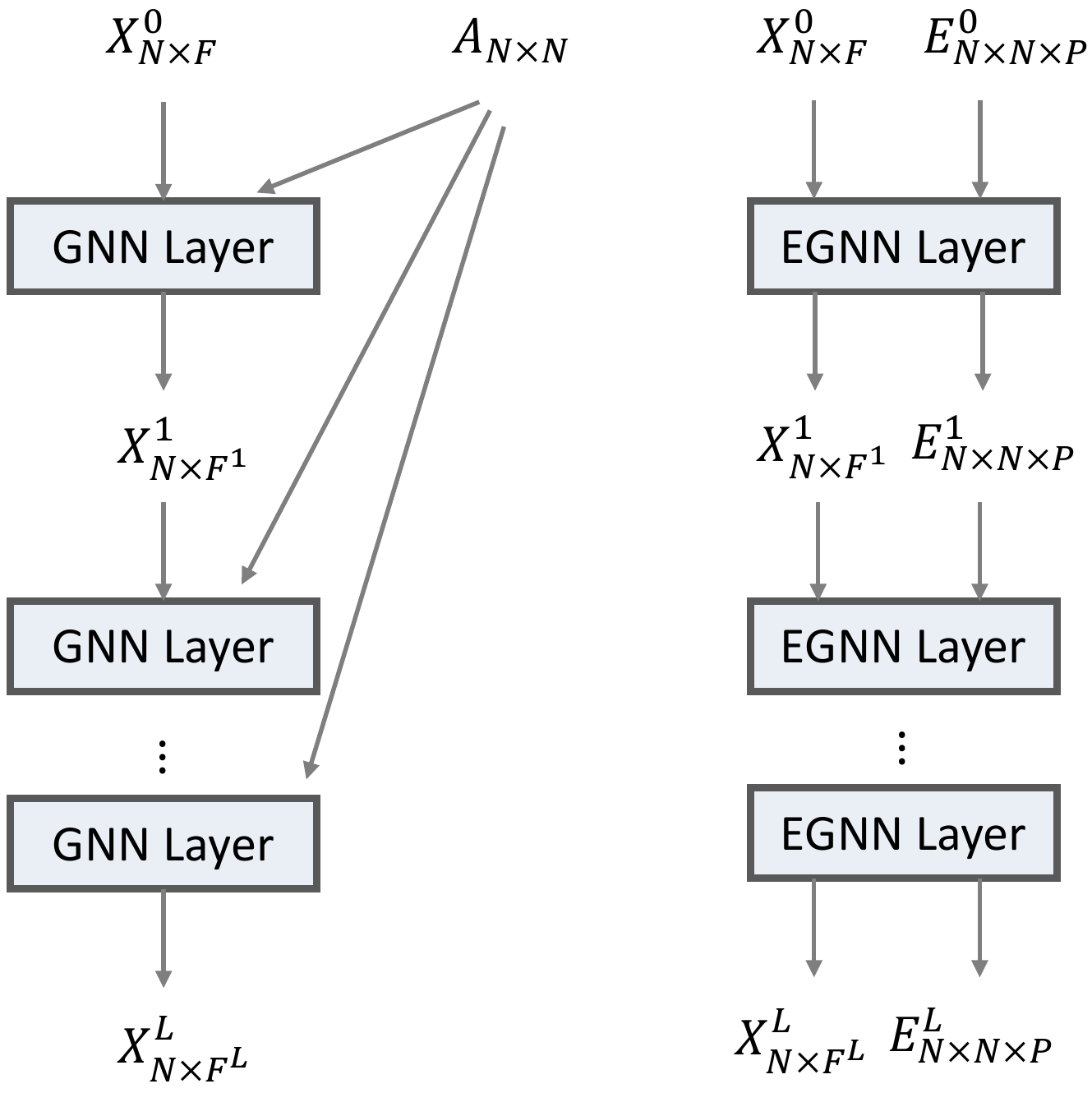}
  \caption{Schematic illustration of the proposed edge enhanced graph
    neural network (EGNN) architecture (right), compared with the
    original graph neural network (GNN) architecture (left). An GNN
    layer could be a GCN layer, or a GAT layer, while a EGNN layer is
    an edge enhanced counterpart of it. EGNN differs from GNN
    structurally in two folds. Firstly, the adjacency matrix
    $\matrixsym{A}$ in GNN is either a binary matrix that indicates
    merely the neighborhood of each node and is used in GAT layers, or
    a positive-valued matrix that has one dimensional edge features
    and is used in GCN layers; in contrast, EGNN uses it with the
    multi-dimensional positive-valued edge features represented as a
    tensor $\matrixsym{E}$ which may exploit multiple attributes
    associated with each edge. Secondly, in GNN the same original
    adjacency matrix $\matrixsym{A}$ is fed to every layer; in
    contrast, the edge features in EGNN are adapted at each layer
    before being fed to next layer.}
  \label{fig:overview}
\end{figure}

Deep neural networks have become one of the most successful machine
learning techniques in recent years. In many important problems, they
achieve state-of-the-art performance, e.g., covolutional neural
networks (CNN) \cite{lecun_gradient-based_1998} in image recognition,
recurrent neural networks (RNN)\cite{elman_finding_1990} and Long
Short Term Memory (LSTM) \cite{hochreiter_long_1997} in natural
language processing, etc. In real world, many problems can be
naturally modeled with graphs rather than conventional tables, grid
type images or time sequences. Generally, a graph contains nodes and
edges, where nodes represent entities in real world, and edges
represent interactions or relationships between entities. For example,
a social network naturally models users as nodes and friendship
relationships as edges. For each node, there is often an asscociated
feature vector describing it, e.g. a user's profile in a socail
network. Similarly, each edge is also often associated with features
depicting relationship strengths or other properties. Due to their
complex structures, a challenge in machine learning on graphs is to
find effective ways to incorporate different sources of information
contained in graphs into models such as neural networks.

Recently, several neural network models have been developed for graph
learning, which obtain better performance than traditional
techniques. Inspired by graph Fourier transform, Defferrard et
al. \cite{defferrard_convolutional_2016} propose a graph covolution
operation as an analogue to standard convolutions used in CNN. Just
like the convolution operation in image spatial domain is equivalent
to multiplication in the frequency domain, covolution operators
defined by polynomials of graph Laplacian is equivalent to filtering
in graph spectral domain. Particularly, by applying Chebyshev
polynomials to graph Laplacian, spatially localized filtering is
obtained. Kipf et al. \cite{kipf_semi-supervised_2017} approximate the
polynomials using a re-normalized first-order adjacency matrix to
obtain comparable results on graph node classification. Those graph
covolutional networks (GCNs)
\cite{defferrard_convolutional_2016}\cite{kipf_semi-supervised_2017}
combine graph node features and graph topological structural
information to make predictions. Velickovic et
al. \cite{velickovic_graph_2018} adopt attention mechanism into graph
learning, and propose a graph attention network (GAT). Unlike GCNs,
which use a fixed or learnable polynomial of Laplacian or adjacency
matrix to aggregate (filter) node information, GAT aggregates node
information by using an attention mechanism on graph
neighborhoods. The essential difference between GAT and GCNs is stark:
In GCNs the weights for aggregating (filtering) neighbor nodes are
defined by the graph topological strcuture, which is independent on
node contents; in contrast, weights in GAT are a function of node
contents due to the attention mechanism. Results on graph node
classification show that the adaptiveness of GAT makes it more
effective to fuse information from node features and graph topological
structures.

One major problem in the current GNN models such as GAT and GCNs is
that edge features are not fully incorporated. In GAT, graph
topological information is injected into the model by forcing the
attention coefficient between two nodes to zero if they are not
connected. Therefore, the edge information used in GAT is only the
indication about whether there is an edge or not,
i.e. connectivities. However, graph edges are often in possession of
rich information like strengths, types, etc. Instead of being a binary
indicator variable, edge features could be continous, e.g. strengths,
or multi-dimensional. GCNs can utilize one-dimensional real value edge
features, \eg edge weights, but the edge features are restricted to be
one-diemansional. Properly addressing this problem is likely to
benefit many graph learning problems. Another problem of GAT and GCNs
is that each GAT or GCN layer filters node features based on the
original adjacency matrix that is given as an input. The original
adjacency matrix is likely to be noisy and not optimal, which will
limit the effectiveness of the filtering operation.

In this paper, we address the above problems by proposing new GNN
models to more adequately exploit edge information, which naturally
enhance current GCNs and GAT models. Our models construct different
formulas from those of GCNs and GAT, so that they are capable of
exploiting multi-dimensional edge features. Also our new models can
exploit one-dimensional edge features more effectively by making them
adaptive across network layers. Moreover, our models leverage doubly
stochastic normalization to augment the GCNs and GAT models that use
ordinary row or symmetric edge normalization. Doubly stochastic
matrices have nice properties that can facilitate the use of edges.

We conduct experiments on several citation network datasets and
molecular datasets. For citation networks, we encode directed edges as
three dimensional edge feature vectors. For molecular datasets,
different atom bond types are naturally encoded as multi-dimensional
edge attributes. By leveraging those multi-dimensional edge features
our methods outperform current state-of-the-art approches. The results
confirm that edge features are important for graph learning, and our
proposed EGAT model effectively incorporates edge features.

As a summary, the novelties of our proposed EGAT model include the
following:
\begin{itemize}
\item A new framework for adequately exploiting multi-dimensional edge
  features. Our new framework is able to incorporate multi-dimensional
  positive-valued edge features. It eliminates the limitation of GAT
  which can handle only binary edge indicators and the limitation of
  GCNs which can handle only one dimensional edge features.
\item Doubly stocahstic edge normalization. We propose to normalize
  edge feature matrices into doubly stochastic matrices which show
  improved performance in denoising \cite{wang_network_2018}.
\item Attention based edge adaptiveness across neural network
  layers. We design a new graph network architecture which can not
  only filter node features but also adapt edge features across
  layers. Leveraging this new architecture, in our model the edge
  features are adaptive to both local contents and the global layers
  when passing through the layers of the network.
\item Multi-dimensional edge features for directed edges. We propose a
  method to encode edge directions as multi-dimensional edge
  features. Therefore, our EGAT can effectively learn on directed
  graph data.
\end{itemize}

The rest of this paper is organized as follows: Section
\ref{sec:related-works} briefly reviews the related works. Details of
the proposed EGNN architecture and two types of proposed EGNN layers
are described in Section \ref{sec:proposed-methods}. Section
\ref{sec:experimental-results} presents the experimental results, and
Section \ref{sec:concl-future-direct} concludes the paper.

\section{Related works}
\label{sec:related-works}

A critical challenge in graph learning is the complex non-Euclidean
structure of graph data. To address this challenge, traditional
machine learning approaches extract graph statistics (e.g. degrees)
\cite{bhagat_node_2011}, kernel functions
\cite{vishwanathan_graph_2010}\cite{shervashidze_weisfeiler-lehman_2011}
or other hand-crafted features which measure local neighborhood
structures. Those methods lack flexibility in that designing sensible
hand-crafted features is time consuming and extensive experiments are
needed to generalize to different tasks or settings. Instead of
extracting structural information or using hand-engineered statistics
as features of the graph, graph representation learning attempts to
embed graphs or graph nodes in a low-dimensional vector space using a
data-driven approach. One kind of embedding approaches are based on
matrix-factorization, e.g. Laplacian Eigenmap (LE)
\cite{belkin_laplacian_2001}, Graph Factorization (GF) algorithm
\cite{ahmed_distributed_2013}, GraRep \cite{cao_grarep:_2015}, and
HOPE \cite{ou_asymmetric_2016}. Another class of approaches focus on
employing a flexible, stochastic measure of node similarity based on
random walks, e.g. DeepWalk \cite{perozzi_deepwalk:_2014}, node2vec
\cite{ahmed_distributed_2013}, LINE \cite{tang_line:_2015}, HARP
\cite{chen_harp:_2018}, etc. There are several limitations in matrix
factorization-based and random walk-based graph learning
approaches. First, the embedding function which maps to
low-dimensional vector space is linear or overly simple so that
complex pattern cannot be captured; Second, they typically do not
incorporate node features; Finally, they are inherently transductive,
for the whole graph structure is required in the training phase.

Recently these limitations in graph learning have been addressed by
adopting new advances in deep learning. Deep learning with neural
networks can represent complex mapping functions and be efficiently
optimized by gradient-descent methods. To embed graph nodes to a
Euclidean space, deep autoencoders are adopted to extract connectivity
patterns from the node similarity matrix or adjacency matrix,
e.g. Deep Neural Graph Representations (DNGR) \cite{cao_deep_2016} and
Structural Deep Network Embeddings (SDNE)
\cite{wang_structural_2016}. Although autoencoder based approaches are
able to capture more complex patterns than matrix factorization based
and random walk based methods, they are still unable to leverage node
features.

With celebrated successes of CNN in image recognition, recently, there
has been an increase interest in adapting convolutions to graph
learning. In \cite{bruna_spectral_2013}, the convolution operation is
defined in the Fourier domain, that is, the spectral space, of the
graph Laplacian. The method is afflicted by two major problems: First,
the eigen decomposition is computationally intensive; second,
filtering in the Fourier domain may result in non-spatially localized
effects. In \cite{henaff_deep_2015}, a parameterization of the Fourier
filter with smooth coefficients is introduced to make the filter
spatially localized. \cite{defferrard_convolutional_2016} proposes to
approximate the filters by using a Chebyshev expansion of the graph
Laplacian, which produces spatially localized filters, and also avoids
computing the eigenvectors of the Laplacian.

Attention mechanisms have been widely employed in many sequence-based
tasks
\cite{bahdanau_neural_2014}\cite{zhou_deep_2017}\cite{kim_semantic_2018}. Compared
with convolution operators, attention machanisms enjoy two benefits:
Firstly, they are able to aggregate any variable sized neighborhood or
sequence; further, the weights for aggregation are functions of the
contents of a neighborhood or sequence. Therefore, they are adaptive
to the contents. \cite{velickovic_graph_2018} adapts an attention
mechanism to graph learning and proposes a graph attention network
(GAT), achieving current state-of-the-art performance on several graph
node classification problems.

\section{Edge feature enhanced graph neural networks}
\label{sec:proposed-methods}

\subsection{Architecture overview}
\label{sec:overview}
Given a graph with $N$ nodes, Let $\matrixsym{X}$ be an $N\times{}F$
matrix representation of the node features of the whole graph. We
denote an element of a matrix or tensor by indices in the
subscript. Specifically, the subscript $\cdot$ is used to select the
whole range (slice) of a dimension. Therefore, $X_{ij}$ will represent
the value of the $j^{th}$ feature of the $i^{th}$
node. $\matrixsym{X}_{i\cdot}\in{}\mathbb{R}^F, i=1,2,\dotsc,N$
represents the $F$ dimensional feature vector of the $i^{th}$
node. Similarly, let $\matrixsym{E}$ be an $N\times{}N\times{}P$
tensor representing the edge feautres of the graph. Then
$\matrixsym{E}_{ij\cdot}\in{}\mathbb{R}^P, i=1,2,\dotsc,N;
j=1,2,\dotsc,N$ represents the $P$-dimensional feature vector of the
edge connecting the $i^{th}$ and $j^{th}$ nodes, and
$\matrixsym{E}_{ijp}$ denotes the $p^{th}$ channel of the edge feature
in $\matrixsym{E}_{ij\cdot}$. Without loss of generality, we set
$\matrixsym{E}_{ij\cdot}=\vectorsym{0}$ to mean that there is no edge
between the $i^{th}$ and $j^{th}$ nodes. Let
$\mathcal{N}_i, i=1,2,\dotsc,N$ be the set of neighboring nodes of
node $i$.

Our proposed network has a multi-layer feedforward architecture. We
use superscript $l$ to denote the output of the $l^{th}$ layer. Then
the inputs to the network are $\matrixsym{X}^0$ and
$\matrixsym{E}^0$. After passing through the first EGAT layer,
$\matrixsym{X}^0$ is filtered to produce an $N\times{}F^1$ new node
feature matrix $\matrixsym{X}^1$. In the mean time, edge features are
adapted to $\matrixsym{E}^1$ that preserves the dimensionality of
$\matrixsym{E}^0$. The adapted $\matrixsym{E}^1$ is fed to the next
layer as edge features. This procedure is repeated for every
subsequent layer. Within each hidden layer, non-linear activations can
be applied to the filtered node features $\matrixsym{X}^l$. The node
features $\matrixsym{X}^L$ can be considered as an embeding of the
graph nodes in an $F^L$-dimensional space. For a node classification
problem, a softmax operator will be applied to each node embedding
vector $\matrixsym{X}_{i\cdot}^L$ along the last dimension. For a
whole-graph prediction (classification or regression) problem, a
pooling layer is applied to the first dimension of $\matrixsym{X}^L$
so that the feature matrix is reduced to a single vector embedding for
the whole graph. Then a fully connected layer is applied to the
vector, whose output could be used as predictions for regression, or
logits for classification. The weights of the network will be trained
with supervision from ground truth labels. Figure \ref{fig:overview}
gives a schematic illustration of the EGNN architecture with a
comparison to the existing GNN architecture. Note that the input edge
features in $\matrixsym{E}^0$ are already pre-normalized. The
normalization method will be described in the next subsection. Two
types of EGNN layers, attention based EGNN (EGNN(A)) layer and
convolution based EGNN (EGNN(C)) layer will also be presented in the
following subsections.

\subsection{Doubly stocahstic normalization of edges}
\label{sec:doubly-stoch-norm}

In graph convolution operations, the edge feature matrices will be
used as filters to multiply the node feature matrix. To avoid
increasing the scale of output features by multiplication, the edge
features need to be normalized. Let $\hat{\matrixsym{E}}$ be the raw
edge features, our normalized features $\matrixsym{E}$ is produced as follows:
\begin{align}
  \label{eq:dsm-1}
  \tilde{\matrixsym{E}}_{ijp} &= \frac{\hat{\matrixsym{E}}_{ijp}}{\sum_{k=1}^N\hat{\matrixsym{E}}_{ikp}}\\
  \label{eq:dsm-2}
  \matrixsym{E}_{ijp} &= \sum_{k=1}^N\frac{\tilde{\matrixsym{E}}_{ikp}\tilde{\matrixsym{E}}_{jkp}}{\sum_{v=1}^N\tilde{\matrixsym{E}}_{vkp}}
\end{align}
Note that all elements in $\hat{\matrixsym{E}}$ are positive. It can
be easily verified that such kind of normalized edge feature tensor
$\matrixsym{E}$ satisfys the following properties:
\begin{align}
  \label{eq:2}
  \matrixsym{E}_{ijp} &\ge 0,\\
  \sum_{i=1}^N\matrixsym{E}_{ijp} &= \sum_{j=1}^N\matrixsym{E}_{ijp} = 1.
\end{align}

In other words, the edge feature matrices
$\matrixsym{E}_{\cdot\cdot{}p}$ for $p=1,2,\cdots,P$ are square
nonnegative real matrices with rows and columns sum to 1. Thus, they
are doubly stochastic matrices, \ie they are both left stochastic and
right stochastic. Mathematically, a stationary finite Markov chain
with a doubly stochastic transition matrix will have a uniform
stationary distribution. Since in a multi-layer graph neural network,
the edge feature matrices will be repeatedly multiplied across layers,
doubly stocahstic normalization could help stablize the process,
compared with the previously used row normalization as in GAT
\cite{velickovic_graph_2018}:
\begin{equation}
  \label{eq:3}
  \matrixsym{E}_{ijp}=\frac{\hat{\matrixsym{E}}_{ijp}}{\sum_{j=1}^N\hat{\matrixsym{E}}_{ijp}}
\end{equation}
or symmetric normalization as in GCN \cite{kipf_semi-supervised_2017}:
\begin{equation}
  \label{eq:4}
  \matrixsym{E}_{ijp}=\frac{\hat{\matrixsym{E}}_{ijp}}{\sqrt{\sum_{i=1}^N\hat{\matrixsym{E}}_{ijp}}\sqrt{\sum_{j=1}^N\hat{\matrixsym{E}}_{ijp}}}
\end{equation}
The effectiveness of doubly stochastic matrix has been recently
demostrated for graph edges denoising \cite{wang_network_2018}.

\subsection{EGNN(A): Attention based EGNN layer}
\label{sec:EGNN-A}
We describe the attention based EGNN layer. The original GAT model
\cite{velickovic_graph_2018} is only able to handle one dimensional
binary edge features, \ie, the attention mechanism is defined on the
node features of the neighborhood, which does not take the real valued
edge features, \eg weights, into account. To address the problem of
multi-dimensional positive real-valued edge features, we propose a new
attention mechanism. In our new attention mechanism, feature vector
$\matrixsym{X}_{i\cdot}^l$ will be aggregated from the feature vectors
of the neighboring nodes of the $i^{th}$ node, i.e.
$\{\matrixsym{X}_j, j\in{}\mathcal{N}_i\}$, by simultaneously
incorporating the corresponding edge features, where $\mathcal{N}_i$
is the indices of neighbors of the $i^{th}$ node. Utilizing the matrix
and tensor notations and the fact that zero valued edge features mean
no edge connections, the aggregation operation is defined as follows:
\begin{equation}
  \label{eq:attention}
  \phantom{.}\matrixsym{X}^l=\sigma\left[\bigparallel_{p=1}^P\left(\matrixsym{\alpha}_{\cdot\cdot{}p}^l(\matrixsym{X}^{l-1}, \matrixsym{E}_{\cdot\cdot{}p}^{l-1})g^l(\matrixsym{X}^{l-1})\right)\right].
\end{equation}
Here $\sigma$ is a non-linear activation; $\matrixsym{\alpha}$ is a
function which produces an $N\times{}N\times{}P$ tensor and
$\matrixsym{\alpha}_{\cdot\cdot{}p}$ is its $p$ channel matrix slice;
$g$ is a transformation which maps the node features from the input
space to the output space, and usually a linear mapping is used:
\begin{equation}
  \label{eq:mapping}
  \phantom{,}g^l(\matrixsym{X}^{l-1}) = \matrixsym{W}^l\matrixsym{X}^{l-1},
\end{equation}
where $\matrixsym{W}^l$ is an $F^l\times{}F^{l-1}$ parameter matrix.

In Eq.\ \eqref{eq:attention}, $\matrixsym{\alpha}^l$ is the so-called
attention coefficients, whose specific entry
$\matrixsym{\alpha}_{ijp}^l$ is a function of
$\matrixsym{X}_{i\cdot}^{l-1}$, $\matrixsym{X}_{j\cdot}^{l-1}$ and
$\matrixsym{E}_{ijp}$, the $p^{th}$ feature channel of the edge
connecting the two nodes. In existing attention mechanisms
\cite{velickovic_graph_2018}, the attention coefficient depends on the
two points $\matrixsym{X}_{i\cdot}$ and $\matrixsym{X}_{j\cdot}$
only. Here we let the attention operation be guided by edge features
of the edge connecting the two nodes, so $\alpha$ depends on edge
features as well. For multiple dimensional edge features, we consider
them as multi-channel signals, and each channel will guide a separate
attention operation.  The results from different channels are combined
by the concatenation operation. For a specific channel of edge
features, our attention function is chosen to be the following:
\begin{align}
  \label{eq:alpha-1}
  \phantom{,}\matrixsym{\alpha}_{\cdot\cdot{}p}^l &= \mathrm{DS}(\hat{\matrixsym{\alpha}}^l_{\cdot\cdot{}p}),\\
  \label{eq:alpha-2}
  \phantom{,}\hat{\matrixsym{\alpha}}_{ijp}^l &= f^l(\matrixsym{X}_{i\cdot}^{l-1}, \matrixsym{X}_{j\cdot}^{l-1}) \matrixsym{E}_{ijp}^{l-1},
\end{align}
where $\mathrm{DS}$ is the doubly stochastic normalization operator
defined in Eqs.\ \eqref{eq:dsm-1} and \eqref{eq:dsm-2}. $f^l$ could
be any ordinary attention function which produces a scalar value from
two input vectors. In this paper, we use a linear function as the
attention function for simplicity:
\begin{equation}
  \label{eq:f}
  \phantom{,}f^l(\matrixsym{X}_{i\cdot}^{l-1}, \matrixsym{X}_{j\cdot}^{l-1}) = \exp\left\{\mathrm{L}\left(\vectorsym{a}^T[\matrixsym{W}\matrixsym{X}_{i\cdot}^{l-1}\Vert{}\matrixsym{W}\matrixsym{X}_{j\cdot}^{l-1}]\right)\right\},
\end{equation}
where $\mathrm{L}$ is the LeakyReLU activation function;
$\matrixsym{W}$ is the same mapping as in \eqref{eq:mapping}; $\Vert$
is the concatenation operation.

The attention coefficients will be used as new edge features for the
next layer, \ie,
\begin{equation}
  \label{eq:e-alpha}
  \matrixsym{E}^l = \matrixsym{\alpha}^l.
\end{equation}
By doing so, EGNN adapts the edge features across the network layers,
which helps capture essential edge features as determined by our new
attention mechanism.

\subsection{EGNN(C): Convolution based EGNN layer}
\label{sec:multi-dimens-edge-2}
Following the fact that graph convolution operation is a special case
of graph attention operation, we derive our EGNN(C) layer from the
formula of EGNN(A) layer. Indeed, the essential difference between
GCN\cite{kipf_semi-supervised_2017} and
GAT\cite{velickovic_graph_2018} is whether we use the attention
coefficients (\ie matrix $\matrixsym{\alpha}$) or adjacency matrix to
aggregate node features. Therefore, we derive EGNN(C) by replacing the
attention coefficient matricies $\matrixsym{\alpha}_{\cdot\cdot{}p}$
with corresponding edge feature matrices
$\matrixsym{E}_{\cdot\cdot{}p}$. The resulting formula for EGNN(C) is
expressed as follows:
\begin{equation}
  \label{eq:5}
  \phantom{,}\matrixsym{X}^l = \sigma\left[\bigparallel_{p=1}^P\left(\matrixsym{E}_{\cdot\cdot{}p}\matrixsym{X}^{l-1}\matrixsym{W}^l\right)\right],
\end{equation}
where the notations have the same meaning as in Section
\ref{sec:EGNN-A}.

\subsection{Edge features for directed graph}
\label{sec:edge-feat-direct}
In real world, many graphs are directed, i.e. each edge has a
direction associated with it. Often times, edge direction contains
important information about the graph. For example, in a citation
network, machine learning papers sometimes cite mathematics papers or
other theoretical papers. However, mathematics papers may seldom cite
machine learning papers. In many previous studies including GCNs and
GAT, edge directions are not considered. In their experiments,
directed graphs such as citation networks are treated as undirected
graphs. In this paper, we show in the experiment part that discarding
edge directions will lose important information. By viewing directions
of edges as a kind of edge features, we encode a directed edge channel
$\matrixsym{E}_{ijp}$ to be
\begin{displaymath}
  \phantom{.}\begin{bmatrix}
    \matrixsym{E}_{ijp} & \matrixsym{E}_{jip} & \matrixsym{E}_{ijp}+\matrixsym{E}_{jip}
  \end{bmatrix}.
\end{displaymath}
Therefore, each directed channel is augmented to three channels. Note
that the three channels define three types of neighborhoods: forward,
backward and undirected. As a result, EGNN will aggregate node
information from these three different types of neighborhoods, which
contains the direction information. Taking the citation network for
instance, EGNN will apply the attention mechanism or convolution
operation on the papers that a specific paper cited, the papers cited
this paper, and the union of the former two. With this kind of edge
features, different patterns in different types of neighborhoods can
be effectively captured.

\section{Experimental results}
\label{sec:experimental-results}
For all the experiments, We implement the algorithms in Python within
the Tensorflow framework \cite{abadi_tensorflow:_2016}. Because the
edge and node features in some datasets are highly sparse, we further
utilize the sparse tensor functionality of Tensorflow to reduce the
memory requirement and computational complexity. Thanks to the sparse
implementation, all the datasets can be efficiently handled by a
Nvidia Tesla K40 graphics card with 12 Gigabyte graphics memory.

\subsection{Citation networks}
\label{sec:citation-networks}
To benchmark the effectiveness of our proposed model, we apply it to
the network node classification problem. Three datasets are tested:
Cora \cite{sen_collective_2008}, Citeseer \cite{sen_collective_2008},
and Pubmed \cite{namata_query-driven_2012}. Some basic statistics
about these datasets are listed in Table \ref{tab:data-summary}.
\begin{table}[ht]
  \centering
  \caption{Summary of citation network datasets}
  \begin{tabular}{l|c|c|c}
    \toprule
     & Cora & Citeseer & Pubmed\\
    \midrule
    \# Nodes & 2708 & 3327 & 19717\\
    \# Edges & 5429 & 4732 & 44338\\
    \# Node Features & 1433 & 1433 & 3703\\
    \# Classes & 7 & 6 & 3\\
    \bottomrule
  \end{tabular}
  \label{tab:data-summary}
\end{table}
All the three datasets are directed graphs, where edge directions
represent the directions of citations. For Cora and Citeseer, node
features contains binary indicators representing the occurrences of
predefined keywords in a paper. For Pubmed, term frequency-inverse
document frequency (TF-IDF) features are employed to describe the
network nodes (\ie papers).

\begin{table*}[tb]
  \centering
  \caption{Classification accuracies on citation networks. Mehods with
    suffix ``-D'' mean no doubly stochastic normalization, thus using
    row normalization in EGNN(A) and using symmetric normalization in
    EGNN(C). Similarly, ``-M'' means ignoring multi-dimensional edge
    features (\ie, using undirected one-dimensional edge features);
    ``-A'' means no adaptiveness across layers; ``*'' means the model
    is trained using weighted loss which takes the class-imbalance of
    training sets into account.}
  \begin{tabular}{l|c|c|c|c|c|c}
    \toprule
    Dataset & \multicolumn{2}{|c|}{Cora} & \multicolumn{2}{c}{CiteSeer} & \multicolumn{2}{|c}{Pubmed}\\
    \hline
    Splitting & Sparse & Dense & Sparse & Dense & Sparse & Dense\\
    \midrule
    \midrule
    GCN    & $72.9\pm{}0.8\%$ & $72.0\pm{}1.2\%$ & $69.2\pm{}0.7\%$ & $75.3\pm{}0.4\%$ & $83.3\pm{}0.4\%$ & $83.4\pm{}0.2\%$\\
    GAT    & $75.5\pm{}1.1\%$ & $79.0\pm{}1.0\%$ & $69.5\pm{}0.5\%$ & $74.9\pm{}0.5\%$ & $83.4\pm{}0.1\%$ & $83.4\pm{}0.2\%$\\
    \midrule
    \midrule
    GCN*   & $82.7\pm{}0.6\%$ & $87.6\pm{}0.6\%$ & $69.3\pm{}0.6\%$ & $76.0\pm{}0.5\%$ & $84.5\pm{}0.2\%$ & $84.3\pm{}0.4\%$\\
    GAT*   & $82.7\pm{0.6}\%$ & $86.6\pm{}0.6\%$ & $69.4\pm{}0.5\%$ & $74.9\pm{}0.8\%$ & $83.1\pm{}0.2\%$ & $82.7\pm{}0.2\%$\\
    \midrule
    \midrule
    EGNN(C)-M & $81.8\pm{}0.5\%$ & $85.1\pm{}0.5\%$ & $\bm{70.6}\pm{}0.3\%$ & $75.0\pm{}0.3\%$ & $84.3\pm{}0.1\%$ & $84.1\pm{}0.1\%$\\
    EGNN(C)-D & $80.2\pm{}0.4\%$ & $86.1\pm{}0.5\%$ & $69.4\pm{}0.3\%$ & $76.8\pm{}0.4\%$ & $\bm{86.2}\pm{}0.2\%$ & $\bm{86.7}\pm{}0.1\%$\\
    EGNN(C)  & $83.0\pm{}0.3\%$ & $\bm{88.8}\pm{}0.3\%$ & $69.5\pm{}0.3\%$ & $76.7\pm{}0.4\%$ & $86.0\pm{}0.1\%$ & $86.0\pm{}0.1\%$\\
    EGNN(A)-D-M  & $76.0\pm{}1.0\%$ & $79.1\pm{}1.0\%$ & $69.5\pm{}0.4\%$ & $74.6\pm{}0.3\%$ & $83.4\pm{}0.1\%$ & $83.6\pm{}0.2\%$\\
    EGNN(A)-A-M  & $80.1\pm{}1.0\%$ & $85.4\pm{}0.5\%$ & $70.1\pm{}0.4\%$ & $74.7\pm{}0.4\%$ & $84.3\pm{}0.2\%$ & $84.2\pm{}0.1\%$\\
    EGNN(A)-A-D  & $81.7\pm{}0.4\%$ & $87.9\pm{}0.4\%$ & $69.4\pm{}0.3\%$ & $75.7\pm{}0.3\%$ & $85.5\pm{}0.1\%$ & $86.0\pm{}0.1\%$\\
    EGNN(A)   & $82.5\pm{}0.3\%$ & $88.4\pm{}0.3\%$ & $69.4\pm{}0.4\%$ & $76.5\pm{}0.3\%$ & $85.7\pm{}0.1\%$ & $\bm{86.7}\pm{}0.1\%$\\
    EGNN(C)-M* & $83.2\pm{}0.3\%$ & $87.4\pm{}0.4\%$ & $70.3\pm{}0.3\%$ & $75.4\pm{}0.5\%$ & $84.1\pm{}0.1\%$ & $84.1\pm{}0.1\%$\\
    EGNN(C)-D* & $82.3\pm{}0.4\%$ & $87.2\pm{}0.4\%$ & $69.4\pm{}0.3\%$ & $\bm{77.1}\pm{}0.4\%$ & $\bm{86.2}\pm{}0.1\%$ & $86.4\pm{}0.3\%$\\
    EGNN(C)*  & $\bm{83.4}\pm{}0.3\%$ & $88.5\pm{}0.4\%$ & $69.5\pm{}0.3\%$ & $76.6\pm{}0.4\%$ & $85.8\pm{}0.1\%$ & $85.6\pm{}0.2\%$\\
    EGNN(A)-D-M* & $82.6\pm{}0.6\%$ & $86.3\pm{}0.9\%$ & $69.4\pm{}0.4\%$ & $74.9\pm{}0.4\%$ & $83.7\pm{}0.2\%$ & $82.8\pm{}0.3\%$\\    
    EGNN(A)-A-M* & $82.7\pm{}0.4\%$ & $87.2\pm{}0.5\%$ & $69.5\pm{}0.3\%$ & $74.5\pm{}0.5\%$ & $83.9\pm{}0.2\%$ & $83.3\pm{}0.2\%$\\
    EGNN(A)-A-D* & $82.8\pm{}0.3\%$ & $87.0\pm{}0.6\%$ & $69.1\pm{}0.3\%$ & $76.3\pm{}0.5\%$ & $85.2\pm{}0.2\%$ & $85.3\pm{}0.3\%$\\
    EGNN(A)*  & $83.1\pm{}0.4\%$ & $88.4\pm{}0.3\%$ & $69.3\pm{}0.3\%$ & $76.3\pm{}0.5\%$ & $85.6\pm{}0.2\%$ & $85.7\pm{}0.2\%$\\
    \bottomrule
  \end{tabular}
  \label{tab:perf-citation}
\end{table*}

The three citation network datasets are also used in
\cite{yang_revisiting_2016} \cite{kipf_semi-supervised_2017}
\cite{velickovic_graph_2018}. However, they all use a pre-processed
version which discards the edge directions. Since our EGNN models
require the edge directions to construct edge features, we use the
original version from \cite{sen_collective_2008} and
\cite{namata_query-driven_2012}. For each of the three datasets, we
split nodes into $3$ subsets for training, validation and testing. Two
splittings were tested. One splitting has $5\%$, $15\%$ and $80\%$
sized subsets for training, validation and test, respectively. Since
it has a small training set, we call it ``sparse'' splitting. Another
splitting has $60\%$, $20\%$ and $20\%$ sized subsets, which is called
``dense'' splitting.

Following the experiment settings of
\cite{kipf_semi-supervised_2017}\cite{velickovic_graph_2018}, we use
two layers of EGNN in all of our experiments for fair
comparison. Throughout the experiments, we use the Adam optimizer
\cite{kingma_adam:_2015} with learning rate $0.005$. An early stopping
strategy with window size of $100$ is adopted for the three citation
networks; i.e. we stop training if the validation loss does not
decrease for $100$ consecutive epochs. We fix the output dimension of
the linear mapping $\matrixsym{W}$ to $64$ for all hidden
layers. Futhermore, we apply dropout \cite{srivastava_dropout:_2014}
with drop rate $0.6$ to both input features and normalized attention
coefficients. $L_2$ regularization with weight decay $0.0005$ is
applied to weights of the model (i.e. $W$ and $a$). Moreover,
exponential linear unit (ELU) \cite{clevert_fast_2016} is employed as
nonlinear activations for hidden layers.

We notice that the class distributions of the training subsets of the
three datasets are not balanced. To test the effects of dataset
imbalanceness, we train each algorithm with two different loss
functions, \ie unweighted and weighted losses, then test performances
of both. The weight of a node belonging to class $k$ is calculated as
\begin{equation}
  \label{eq:6}
  \frac{\sum_{k=1}^Kn_k}{Kn_k},
\end{equation}
where $K$ and $n_k$ are the numbers of classes and nodes belonging to
the $k^{th}$ class in the training subset, respectively. Basically,
nodes in a minority class are given larger weights than a majority
class, and thus are penalized more in the loss.

\begin{figure*}[ht!]
  \centering
  \begin{tabular}{ccc}
    \includegraphics[scale=0.25]{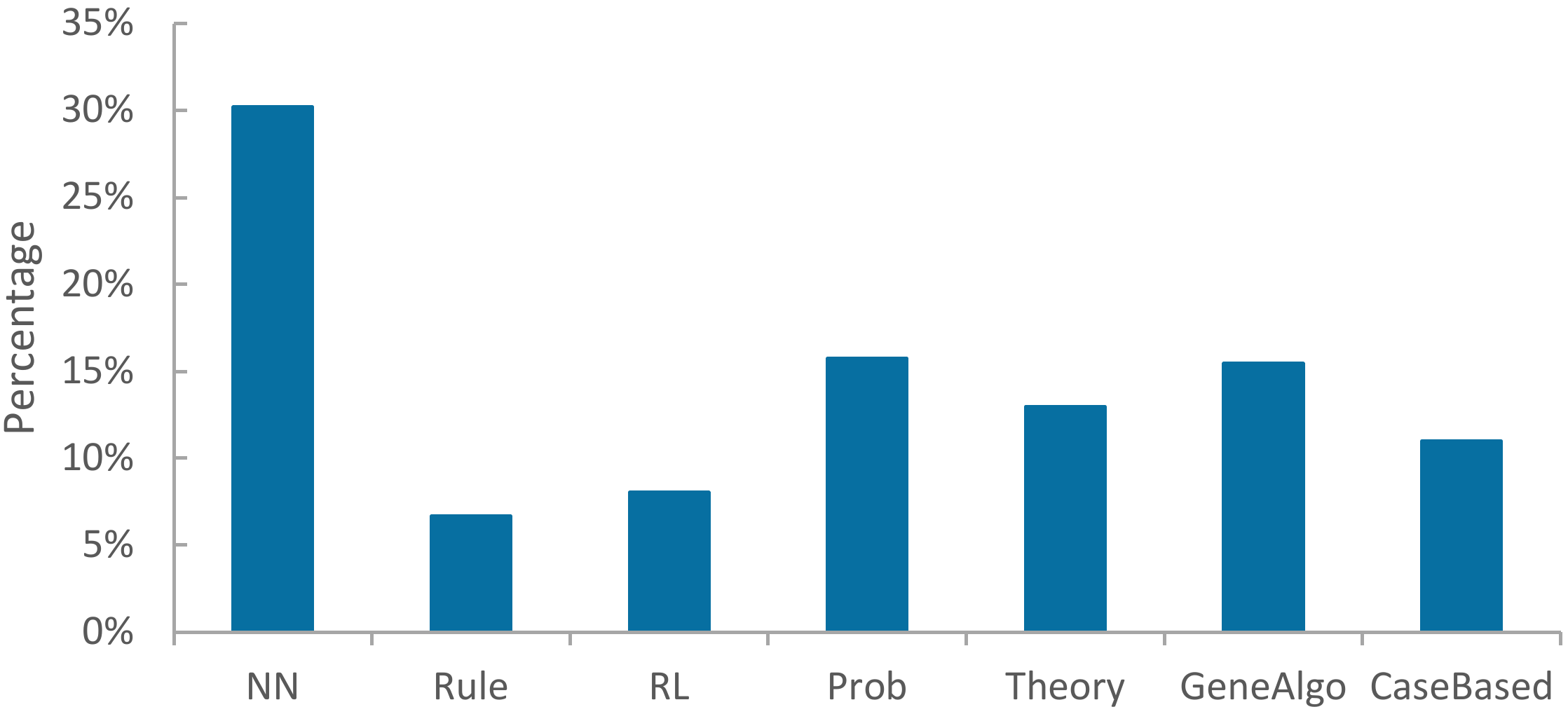}
    & \includegraphics[scale=0.3]{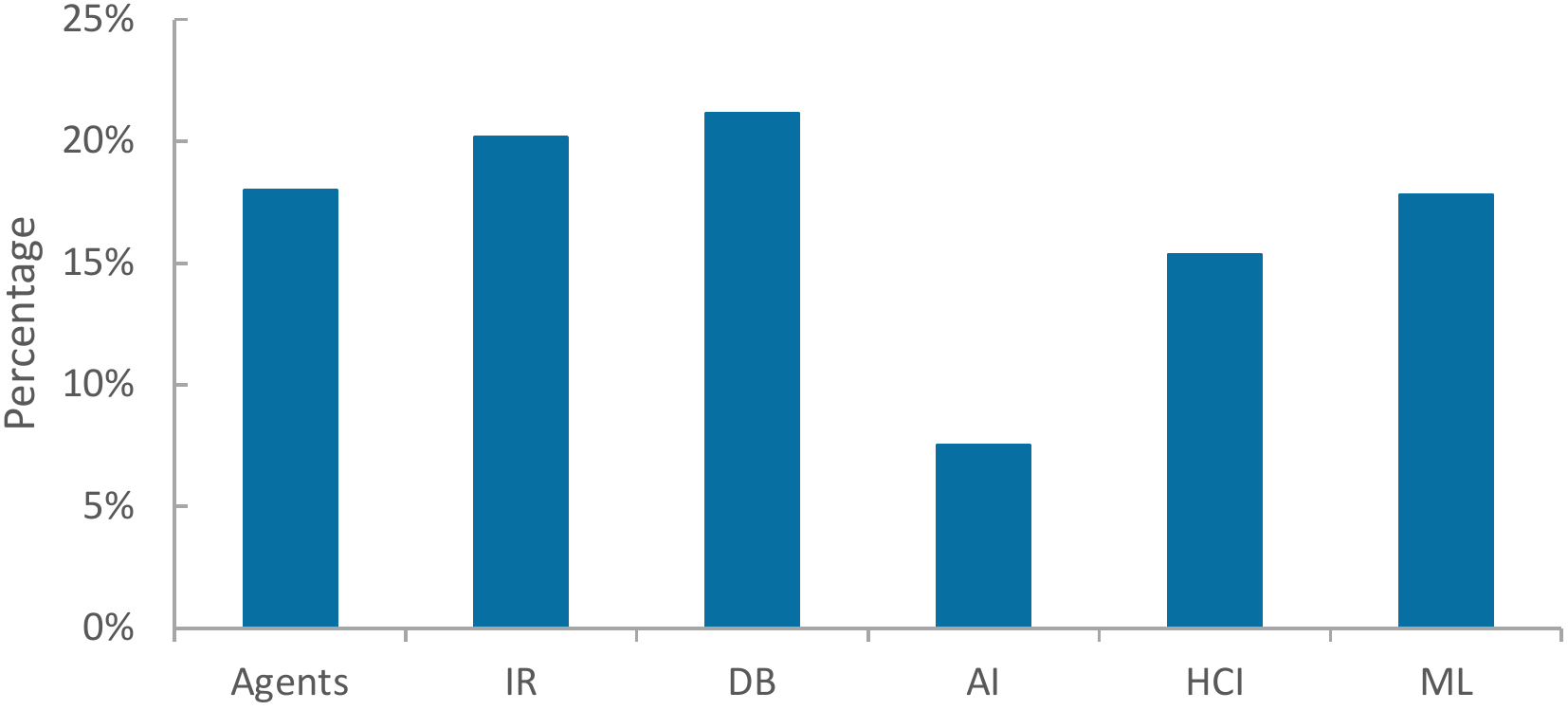}
    & \includegraphics[scale=0.3]{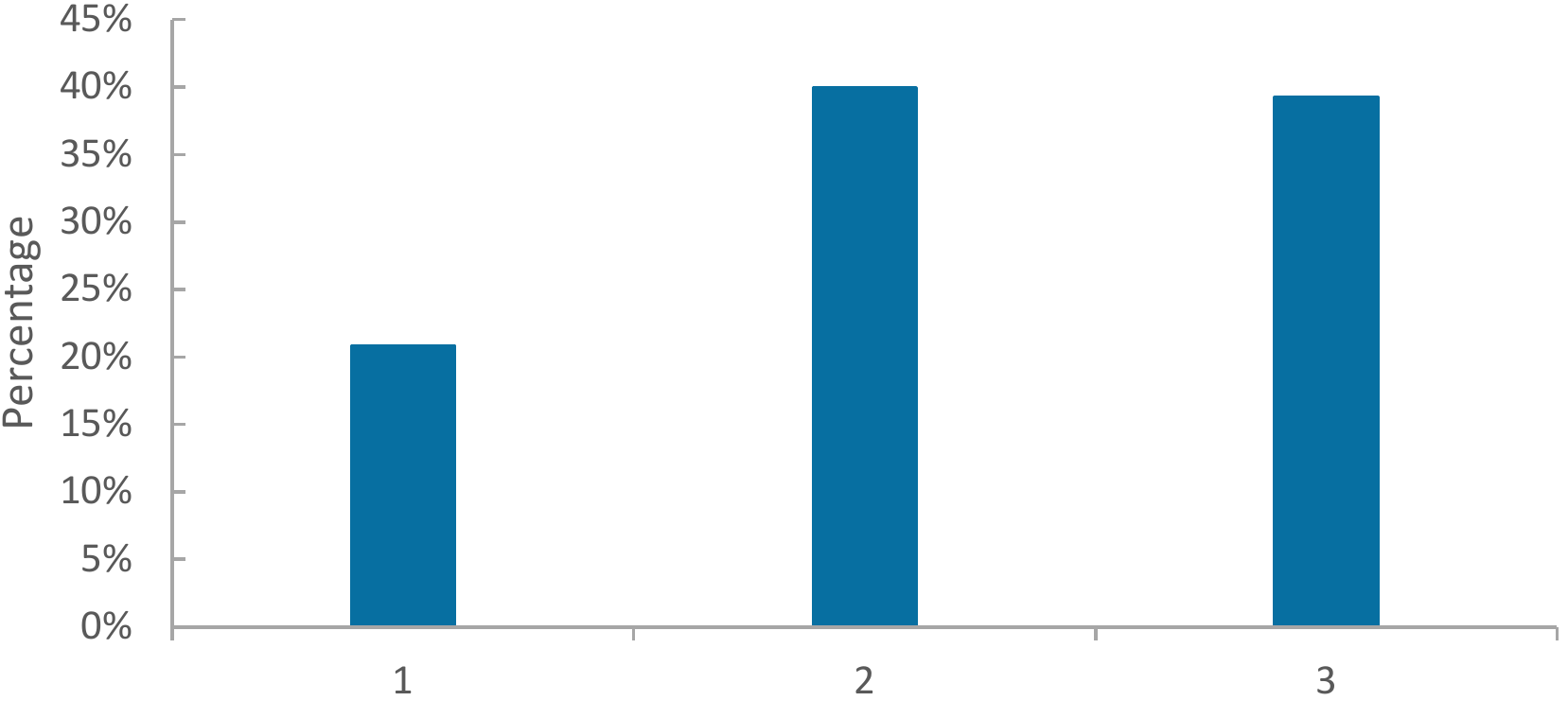}\\
    (a) Cora
    & (b) Citeseer
    & (c) Pubmed
  \end{tabular}
  \caption{Node class distribution of the training subsets of the
    three citation networks. The Cora dataset is more imbalanced than
    the other two.}
  \label{fig:class-dist}
\end{figure*}

The baseline methods we used are GCN \cite{kipf_semi-supervised_2017}
and GAT \cite{velickovic_graph_2018}. To further investigate the
effectivenesses of each components, \ie doubly stochastic
normalization, multi-dimensional edge features and edge adaptiveness,
we also test different versions of EGNN(A) and EGNN(C) that keep only
one component and discard the others. The performances are recorded
for abalation study. Totally, $9$ models are tested:
\begin{itemize}
\item GCN: baseline as described in \cite{kipf_semi-supervised_2017}.
\item GAT: baseline as described in \cite{velickovic_graph_2018}.
\item EGNN(C)-M: EGNN(C) variant which ignores multi-dimensional edge
  features, \ie, it treats directed edges as undirected ones. Note
  that the doubly stocahstic normalization component is kept.
\item EGNN(C)-D: EGNN(C) variant without doubly stochastic normalization.
\item EGNN(C): Full EGNN(C) with doubly stochastic normalization and
  multi-dimensional edge features.
\item EGNN(A)-D-M: EGNN(A) variant without doubly stochastic
  normalization as well as without multi-dimensional edge features.
\item EGNN(A)-A-M: EGNN(A) variant without edge adaptiveness across
  layers as well as multi-dimensional edge features.
\item EGNN(A)-A-D: EGNN(A) variant without edge adaptiveness across
  layers as well as doubly stochastic normalization.
\item EGNN(A): Full EGNN(A) with all functionalities.
\end{itemize}
Note that each algorithm has both a weighted loss version and
unweighted loss version.

We run each version of the algorithms $20$ times, and record the mean
and standard deviation of the classification accuracies, which are
listed in Table \ref{tab:perf-citation}. From the table, we can
observe serval interesting phenomena which warrant further
investigations:
\begin{itemize}
\item Overall, almost all EGNN variants outperform their corresponding
  baselines, which indicates that all the three components incorporate
  useful information for classification. Particularly,
  multi-dimensional edge features and doubly stochastic normalization
  improve more than edge adaptiveness.
\item The two baselines fail on both the sparse and dense splittings of
  the Cora dataset. This is caused by the class imbalance of the Cora
  dataset. We illustrate the class distributions of the three datasets
  in Figure \ref{fig:class-dist}. From the distributions, we can see
  that Cora is more imbalanced than Citeseer and Pubmed.
\item On the Cora dataset, the baselines with weighted loss perform
  normal. Again, this indicates that their failures are caused by the
  class imbalance.
\item Our proposed methods are highly resistant to class
  imbalance. Without weighted training, our framework obtain high
  accuracies on the Cora dataset.
\item Weighted training does not always improve performance,
  especially on less imbalanced datasets, e.g. Pubmed. This indicates
  that simply weighting the nodes is not sufficient to fully solve the
  class imbalance problem. Therefore, more sophisticated methods need
  to be designed to address this problem.
\item Performances on dense splittings are consistently higher than on
  sparse splitting. It is not unexpected because more training data
  gives an algorithm more information to tune parameters.
\item Either EGNN(C)-M* or EGNN(C)-D* is close to or a little bit
  worse than GCN* on the dense splitting of the Cora dataset. However,
  EGNN(C)* is considerably better than GCN*. This interesting
  phenomena indicates doubly stochastic normalization and
  multi-dimensional edge feature may not work well individually on
  some datasets, but can improve performance considerably if combined.
\end{itemize}

\subsection{Molecular analysis}
\label{sec:molecular-analysis}

\begin{table*}[t]
  \caption{Performance on molecular datasets}
  \centering
  \begin{tabular}[h]{l|c|c|c|c|c|c}
    \toprule
    \multirow{3}{*}{Dataset} & \multicolumn{2}{c}{Tox21 (AUC)} & \multicolumn{2}{|c|}{Lipo (RMSE)} & \multicolumn{2}{c}{Freesolv (RMSE)}\\
    & Validation & Test & Validation & Test & Validation & Test\\
    \midrule
    \midrule
    RF & $0.78\pm{}0.01$ & $0.75\pm{}0.03$ & $0.87\pm{}0.02$ & $0.86\pm{}0.04$ & $1.98\pm{}0.07$ & $1.62\pm{}0.14$\\
    Weave & $0.79\pm{}0.02$ & $0.80\pm{}0.02$ & $0.88\pm{}0.06$ & $0.89\pm{}0.04$ & $1.35\pm{}0.22$ & $1.37\pm{}0.14$\\
    \midrule
    \midrule
    EGNN(C) & $\bm{0.82}\pm{}0.01$ & $\bm{0.82}\pm{}0.01$ & $0.80\pm{}0.02$ & $\bm{0.75}\pm{}0.01$ & $\bm{1.07}\pm{}0.08$ & $1.09\pm{}0.08$\\
    EGNN(A) & $\bm{0.82}\pm{}0.01$ & $0.81\pm{}0.01$ & $\bm{0.79}\pm{}0.02$ & $\bm{0.75}\pm{}0.01$ & $1.09\pm{}0.12$ & $\bm{1.01}\pm{}0.12$\\
    \bottomrule
  \end{tabular}
  \label{tab:perf-molecular}
\end{table*}

One promising application of graph learning is molecular analysis. A
molecular can be represented as a graph, where each atom is a node,
and chemical bonds are edges. Unlike citation network analysis in
Section \ref{sec:citation-networks}, the problem here is whole-graph
prediction, either classification or regression. For example, given a
graph representation of a molecular, the goal might be to classify it
as toxic or not, or to predict the soluability (regression). In other
words, we need to predict one value for the whole graph, rather than
one value for a graph node. Usually, for each chemical bond, there are
several attributes associated with it, \eg, Atom Pair Type, Bond
Order, Ring Status, etc. Therefore, the graphs intrinsically contain
multi-dimensional edge features.

Three datasets (Tox21, Lipophilicity and Freesolv) are used to test
our algorithms. Tox21 contains $7831$ enviromental compounds and
drugs. Each compound is associated with $12$ labels, \eg androgen
receptor, estrogen receptor, and mitochondrial membrane potential,
which defines a multi-label classification problem. Lipophilicity
contains $4200$ compounds. The goal is to predict compound
soluability, which is a regression task. Freesolv includes a set of
$642$ neutral molecules, which similarly defines a regression
task. For all the three datasets, compounds are converted to
graphs. For all the three datasets, nodes are described by $25$-d
feature vectors. The dimensionality of edge feature vectors are $42$,
$21$ and $25$ for Tox21, Lipo, and Freesolv, respectively.

For both EGNN(A) and EGNN(C), we implement a network containing $2$
graph processing layers, a global max-pooling layer, and a fully
connected layer. For each graph processing layer, the output
dimensions of the linear mapping $g$ are fixed to be $16$. For Tox21,
sigmoid cross entropy losses are applied to the output logits of the
fully connected layer. For Lipo and Freesolv, mean squared error
losses are employed. The networks are trained by Adam optimizer
\cite{kingma_adam:_2015} with learning rate $0.0005$. An early
stopping strategy with window size of $200$ is adopted. $L_2$
regularization with weight decay $0.0001$ is applied to parameters of
the models except bias parameters. Moreover, exponential linear unit
(ELU) \cite{clevert_fast_2016} is employed as nonlinear activations
for hidden layers.

Our methods are compared with two baseline models which are shown in
MoleculeNet \cite{wu_moleculenet:_2018}: Random Forest and
Weave. Random Forest is a traditional learning algorithm which is
widely applied to various problems. Weave model
\cite{kearnes_molecular_2016} is similar to graph convolution but
specifically designed for molecular analysis.

All the three datasets are split into training, validation and test
subsets sized $80\%$, $10\%$ and $10\%$, respectively. We run our
models $5$ times, and record the means and standard deviations of
performance scores. For classification task (\ie, Tox21), Area Under
Curve (AUC) scores of the receiver operating characteristic (ROC)
curve is recorded. Since it is a multi-label classification problem,
we record the AUCs of each class and take the average value as the
final score. For regression (\ie, Lipo and Freesolv), root mean square
error (RMSE) are recorded. We list the scores in Table
\ref{tab:perf-molecular}. The results show that our EGNN(C) and
EGNN(A) outperform the two baselines with considerable margins. On the
Tox21 dataset, the AUC scores are improved by more than $0.2$ compared
with the Weave model. For the two regression tasks, RMSEs are improved
by about $0.1$ and $0.3$ on the Lipo and Freesolv datasets,
respectively. On the other hand, the scores of EGNN(C) and EGNN(A) are
very close on the three datasets.

\section{Conclusions}
\label{sec:concl-future-direct}

In this paper, we propose a new framework to address the existing
problems in the current state-of-the-art graph neural network
models. Specifically, we propose a new attention mechanism by
generalizing the current graph attention mechansim used in GAT to
incorporate multi-dimensional real-valued edge features. Then, based
on the proposed new attention mechanism, we propose a new graph neural
network architecture that adapts edge features across neural network
layers. Our framework admits a formula that allows for extending
convolutions to handle multi-dimensional edge features. Moreover, we
propose to use doubly stochastic normalization, as opposed to the
ordinary row normalization or symmetric normalization used in the
existing graph neural network models. Finally, we propose a method to
design multi-dimensional edge features for directed edges so that our
model is able to effectively handle directed graphs. Extensive
experiments are conducted on three citation network datasets for graph
node classification evaluation, and on three molecular datasets to
test the performance on whole graph classification and regression
tasks. Experimental results show that our new framework outperforms
current state-of-the-art models such as GCN and GAT consistently and
significantly on all the datasets. Detailed abalation study also show
the effectiveness of each individual component in our model.

{\small
\bibliographystyle{ieee}
\bibliography{ref}
}

\end{document}